# Improvement of Color Image Analysis Using a New Hybrid Face Recognition Algorithm based on Discrete Wavelets and Chebyshev Polynomials

*Hassan M. Muhialdeen*
*Department of Computer Engineering*
*Aliraqia University*
*22Sabaabkar, Adamia, Baghdad, Iraq*
*muhialdeen.hassan@aliraqia.edu.iq*
*https://orcid.org/0000-0002-9177-8621*

*Asma A. Abdulrahman*
*Department of Applied Sciences*
*University of Technology*
*52Alsena street,Baghdad, Iraq*
*Asma.A.Abdulrahman@uotechnology.edu.iq*
*http://orcid.org/0000-0001-6034-8859*

*Jabbar Abed Eleiwy*
*Department of Applied Sciences*
*University of Technology*
*52Alsena street Baghdad, Iraq*
*jabar.a.eleiwy@uotechnology.edu.iq*
*https://orcid.org/0000-0002-7442-1719*

*Fouad S. Tahir*
*Department of Applied Sciences*
*University of Technology, Iraq*
*52Alsena street, Baghdad, Iraq*
*fouad.s.taher@uotechnology.edu.iq*
*http://orcid.org/0000-0002-3121-201X*

*Yurii Khlaponin*
*Corresponding author*
*Department of Cybersecurity and Computer Engineering*
*Kyiv National University of Construction and Architecture*
*31 Povitroflotskyi ave, Kyiv, Ukraine, 03037*
*y.khlaponin@knuba.edu.ua*
*https://orcid.org/0000-0002-9287-0817*

*Mushtaq Talib Al-Sharify*
*Department of Communication Technical Engineering, Al-Farahidi University, Baghdad, Iraq*
*https://orcid.org/0000-0002-9818-3612*

alsharify@uoalfarahidi.edu.iq

**Abstract:**
This work is unique in the use of discrete wavelets that were built from or derived from Chebyshev polynomials of the second and third kind, filter the Discrete Second Chebyshev Wavelets Transform (DSCWT), and derive two effective filters. The Filter Discrete Third Chebyshev Wavelets Transform (FDTCWT) is used in the process of analyzing color images and removing noise and impurities that accompany the image, as well as because of the large amount of data that makes up the image as it is taken. These data are massive, making it difficult to deal with each other during transmission. However to address this issue, the image compression technique is used, with the image not losing information due to the readings that were obtained, and the results were satisfactory. Mean Square Error (MSE), Peak Signal Noise Ratio (PSNR), Bit Per Pixel (BPP), and Compression Ratio (CR) Coronavirus is the initial treatment, while the processing stage is done with network training for Convolutional Neural Networks (CNN) with Discrete Second Chebeshev Wavelets Convolutional Neural Network (DSCWCNN) and Discrete Third Chebeshev Wavelets Convolutional Neural Network (DTCWCNN) to create an efficient algorithm for face recognition, and the best results were achieved in accuracy and in the least amount of time. Two samples of color images that were made or implemented were used. The proposed theory was obtained with fast and good results; the results are evident shown in the tables below.

**Key words:** Image processing, Artificial Intelligent (AI), Discrete Chebyshev wavelets Transform (DCWT), Convolutional Neural Network (CNN)

**Introduction**

Artificial intelligence (AI) is the reason for meeting the demands of modern life in terms of health and economy in the fields of image processing [1], disease analysis and diagnosis [2], the development of treatments for these diseases [3], as well as in the fields of surgery [4], and disease follow-up [5]. AI is the mechanism for programming human thought. what artificial intelligence has accomplished Even in veterinary medicine, artificial intelligence was used to overcome the challenges that life faced in previous eras. With the training of convolutional neural networks (CNN) with the classification ResNet50, and reaching an accuracy of 99.8%, with an error rate of 0.005 [6-8], deep learning found its way into a variety of fields, including security and medical fields, such as face recognition as well as recognition, and medical fields, such as brain tumors. In the recent past, after the Coronavirus, which causes lung inflammation and death, where an accuracy of 97.5% was attained by providing an approach in medical images classification to diagnose the disease [9], Deep learning and a convolutional neural network are used to diagnose fractures in human bones. CNN was trained for a lengthy period of time in image recognition, but only with a low error rate in image analysis [10]. Typically, x-rays are used by doctors to identify bone fractures. Due to the lengthy nature of this detection, deep learning was required to train the neural network.

Performance evaluation revealed an accuracy of 92.44%, which was twice as good as the accuracy obtained without deep learning [11], [12], 84.7%, and 86% [13]. Many tasks are simplified by deep learning and convolutional neural networks. Orthopedic injuries were identified as the cause. With CrackNet, the fast R-CNN convolutional neural network was found [14]. Recently, convolutional neural networks were used with watermark technology to protect digital images. The host image and the digital image were processed, and when the watermark was compromised, the convolutional neural network was projected into 13 layers to extract it. The mean square error (MSE) and the absolute error (MAE) were computed [15]. Wave transformations with watermarks were utilized in the convolutional neural networks (CNNs) procedure, where 11 CNN layers were used, to reduce the complexity of retrieving the watermark without losing the quality of the digital images [16]. Convolutional neural networks are able to detect faces and extract information about the individual whose face is being identified from visual data. This is about a 10% increase in recognition accuracy achieved by network training. In order to enhance the identification of wrfracturesture, Frat Hardalaç et al. used deep learning on wrist X-ray imaging data at the Gazi University Hospital. With an average accuracy (AP50) score of 0.8639, deep learning models were developed [18].

A convolutional neural network and deep learning were used to identify gender categorization. It took four layers to get an accuracy of 8.759%. This method has applications in forensic medicine, where it integrates face measuring for images with deep learning at a pace of more than 3700 images per second with an accuracy of 89%. [19-20]. The work done by S. Laith and others using deep learning in face detection in the MATLAB programme by training the convolutional neural network and achieving 100% accuracy, [21-22], produced the most astounding results. The best results were obtained from these wavelets: Discrete First Chebyshev Wavelet Transform (DCFWT), used in image processing for including noise lifting and compression, and Discrete Hermite Wavelet Transform (DH). Asma and others built new discrete wavelets in addition to those previously used, which are the fundamental discrete wavelets such as Haar and others, which were used in image processing for removing noise from the colour image and compressing the image. The same wavelets as Fouad in order to detect faces and develop a quick technique for wavelet convolutional neural network W-CNN training, these distinct wavelets [DCFWT] were employed [25–27]. Shaymaa The quick technique was developed to achieve the necessary precision, and facial emotions were recognised using wavelets (DCHWT) and convolved using a neural network known as the Chebyshev wavelet convolutional neural network (CHWCNN) [28–31]. This work used a novel filter produced from discrete wavelets to address the issues of impurity, noise, and picture compression. The convolutional neural networks were then trained using the discrete second Chebyshev wavelets transform (DSCWT), discrete third Chebyshev wavelets transform (DTCWT), and filter discrete second Chebyshev wavelets transform (FDTCWT). Discrete Third Chebyshev Wavelet Convolution Neural

Network (DTCWCNN) and Discrete Second Chebyshev Wavelet Convolution Neural Network (DSCWCNN). The best outcomes were obtained, demonstrating with two samples of colored images the effectiveness of the suggested hypothesis. Face recognition has been carried out on the two images for which the proposed algorithm was used in this study, and the images were then presented to a convolutional neural network with separate wavelets of the second and third Chebyshev figures after the most crucial quality criteria for the colour image were calculated and the results were compared before and after noise removal. Using DSCWCNN with a duration of 1 min and 1 sec and DTCWCNN with a time of one min and Nineteen sec, respectively, and representing the two samples as one, experiments will be conducted on them and their accuracy will be 98.60% and 98.92%, respectively.

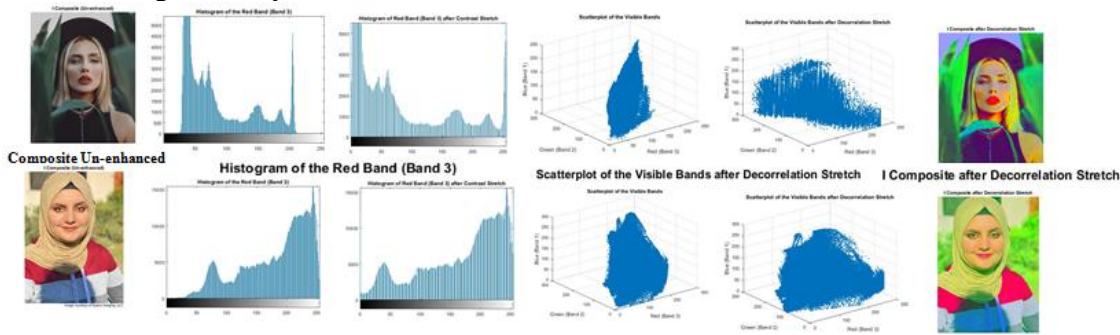

Figure 1. represents the two samples

## 2. Methodology

This section presents the most important equations of Discrete wavelet Transformation (DWT) derived from Second and Third Chebyshev polynomials because of their importance in dealing with image processing in the field of Multi Resolution Analyses (MRA) due to the orthogonally and convergent characteristic of these wavelets, this wavelet is Discrete Second and Third Chebyshev Wavelet Transformation (DSCWT)and (DTCWT) after which the convolutional neural network is trained with the new wavelets, Discrete Second Chebyshev Wavelet Convolution Neural Network (DSCWCNN), and Discrete Third Chebyshev Wavelet Convolution Neural Network (DTCWCNN), Figure.2 shows the work steps and the proposed theory.

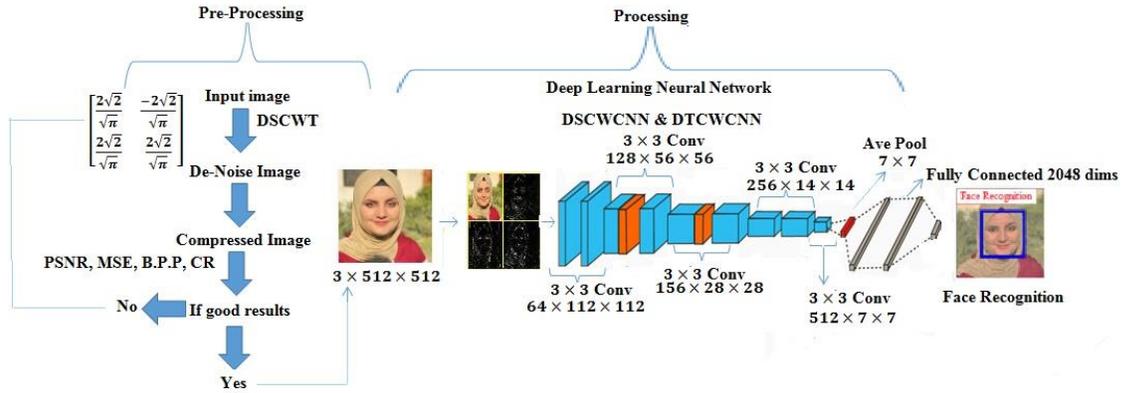

Figure 2. A diagram showing the steps of the proposed work

The same above processing reached the diagram with Discrete Third Chebyshev Wavelet Convolution Neural Network (DTCWCNN).

## 2.1. Pre-Processing

Computer vision in image processing is the use of wavelets, where the noise is removed from the medical image and the image is compressed, because the wavelet transform has wonderful advantages because it transforms into the frequency and time domain, and it is also characterized by flexibility and simplicity of its arithmetic operations for the wavelets Haar, Meyer, Morlettich known.

The set of functions in data analysis that are carried out by the wavelets that are considered the basis and which are called the mother wavelets and are better in their work than the sine and the perfect sine with respect to the frequency domain and time, through the sudden high and low frequencies, wave coefficients will be created that are useful in determining changes in the signals that are directly dependent on these changes, the accuracy of the wavelets.

In the past ten years, waves have taken their way in identifying the face because they have the ability to show or clarify the characteristics of the face.

Haar Discrete Wavelet Transform was used. In this work, DiscreteSecondChebyshev wavelets Transform (DSCWT) and Discrete Third Chebyshev wavelets Transform (DTCWT) on the interval $[0,1)$ in equations (1) and (2)

$$W_{r,s}^2(t) = \begin{cases} 2^{\frac{k}{2}} \widetilde{W}_s(2^k t - 2r + 1) & \frac{r-1}{2^{k-1}} \leq t < \frac{r}{2^{k-1}} \\ 0 & o.w \end{cases} \quad (1)$$

$$\text{where} \qquad \widetilde{W}_s(t) = \sqrt{\frac{2}{\pi}} W_s(t)$$

$$W_{r,s}^3(t) = \begin{cases} 2^{\frac{k}{2}} & \widetilde{W}_s(2^{k+1}t - 2r + 1) \frac{r-1}{2^k} \leq t < \frac{r}{2^k} (2) \\ 0 & \text{oterwise} \end{cases}$$

where $\widetilde{W}_s(t) = \frac{1}{\sqrt{\pi}} W_s(t)$

Filter Discrete Second Chebyshev wavelets Transform (FDSCWT)= $\begin{bmatrix} \frac{2\sqrt{2}}{\sqrt{\pi}} & \frac{-2\sqrt{2}}{\sqrt{\pi}} \\ \frac{2\sqrt{2}}{\sqrt{\pi}} & \frac{2\sqrt{2}}{\sqrt{\pi}} \end{bmatrix}$

Filter Discrete Third Chebyshev wavelets Transform (FDTCWT)= $\begin{bmatrix} \frac{\sqrt{2}}{\sqrt{\pi}} & \frac{-\sqrt{2}}{\sqrt{\pi}} \\ \frac{\sqrt{2}}{\sqrt{\pi}} & \frac{\sqrt{2}}{\sqrt{\pi}} \end{bmatrix}$

The two separate waves will be tested in denoising the color image and compressing the image in the preliminary stage, Second and Third Kind Chebyshev (DWT), will be used, which is considered one of the wavelets that are characterized by their work with a frequency domain, The detected details of the face with the separated wavelets with less execution time and because of the addition and subtraction using the newSecond and Third Kind Chebyshev (DWT). The first step is to use the proposed transformation Second and Third Kind Chebyshev (DWT) on the input image two dimension. The image analysis process begins Multi Resolution Analyses (MRA), In the figure (3) the entered image parameters will be analysed into the approximateConfessions and the detail Confessions so that the image is divided into four parts. The first part is the approximating part and is symbolized by Low Low (LL), while the remaining three parts include detail Confessions to be High Low (HL) Low High (LH) and High High (HH) respectively in the first level, and the same process is repeated in the second level so that the first quadrant is taken so that the filter is passed on it so that the image is divided into four other parts and so on and back to the original image after The analysis after taking the inverse of the discrete wavelets.

while preserving the image properties through the feature carried by the wavelets, which is the transition of small waves to large waves while reducing the parameters while preserving the features of the original image after taking the inverse of the Inverse Second and Third Kind Chebyshev (DWT) to be reconstructed for the input image

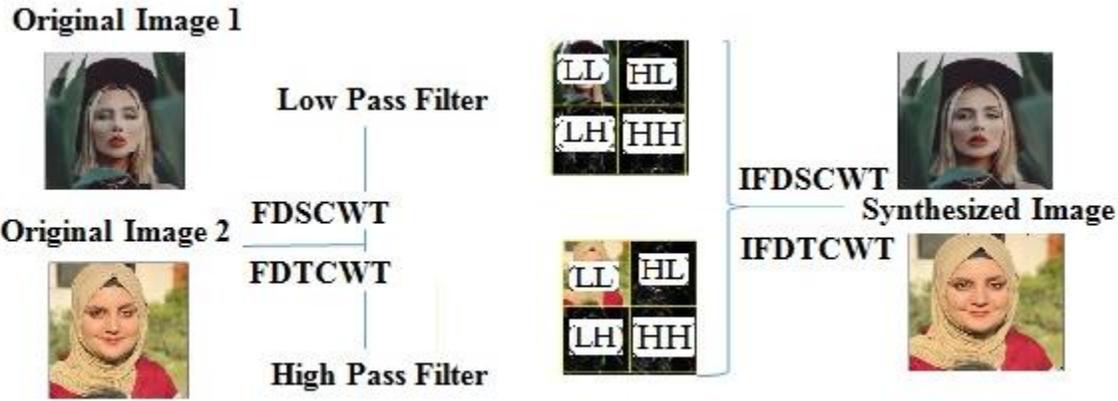

Figure.3. image analysis process with Filter Discrete Second Chebyshev wavelets Transform (FDSCWT)

Same figure with FDTCWT. Figure (4 ) shows the complete process, while removing noise from the color image by using the new filter for the new wavelets Second and Third Kind Chebyshev (DWT).

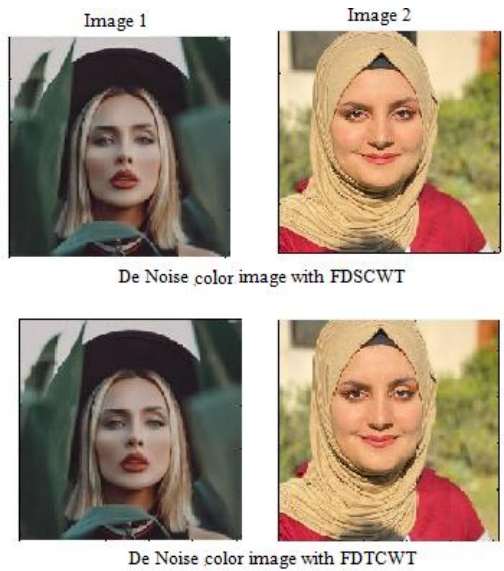

Figure 4. shows the complete process, while removing noise

As for the process of compressing the image using the proposed technique Second and Third Kind Chebyshev (DWT) with Set Partitioning in Hierarchical Trees (SPIHT) because of the characteristic that this technique bears, it is the orthogonal property that works to reduce the space occupied by the color image data by reducing the pixels of the image while increasing the image compression, The most important criteria for image quality are measured the Mean Square Error (MSE), Peak Signal to Noise Ratio (PSNR), Bit Per Pixel (BPP), and Compression Ratio (CR).

TablesShows the above operation with results and comparison of pressure with FDSCWT and FDTCW

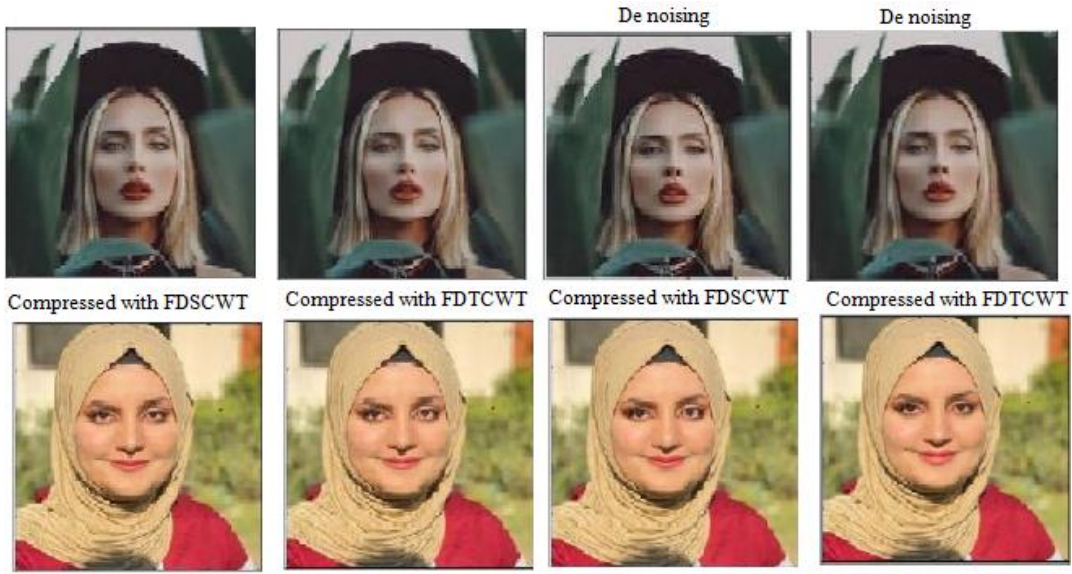

Figure 5. The compression before and after de nosing Image1and Image 2 used FDSCWT and FDTCWT

Table 1: The compression before de nosing Image1 used FDSCWT and FDTCWT

| Iterations | FDSCWT & FDTCWT | | | |
|---|---|---|---|---|
| | MSE | PSNR | BPP | CR |
| 1 | 10570 | 7.889 | 0.008 | 0.03 |
| 2 | 3372 | 12.85 | 0.008 | 0.03 |
| 3 | 3372 | 12.85 | 0.008 | 0.03 |
| 4 | 2852 | 13.58 | 0.008 | 0.03 |
| 5 | 1811 | 15.58 | 0.015 | 0.05 |
| 6 | 1366 | 16.78 | 0.016 | 0.07 |
| 7 | 863.7 | 18.77 | 0.039 | 0.16 |
| 8 | 555.1 | 20.69 | 0.092 | 0.38 |
| 9 | 256.9 | 24.03 | 0.246 | 1.03 |
| 10 | 117.2 | 27.44 | 0.538 | 2.24 |
| 11 | 52.1 | 30.96 | 1.074 | 4.48 |
| 12 | 21.27 | 34.85 | 1.961 | 8.17 |
| 13 | 8.756 | 38.71 | 3.253 | 13.55 |
| 14 | 3.947 | 42.17 | 5.155 | 21.48 |
| 15 | 2.28 | 44.55 | 7.783 | 32.42 |
| 16 | 1.802 | 45.57 | 11.056 | 46.07 |
| 17 | 1.802 | 45.57 | 11.056 | 46.07 |

**Table 2: The compression before de nosing Image2 used FDSCWT and FDTCWT**

| Iterations | FDSCWT & FDTCWT | | | |
|---|---|---|---|---|
| | MSE | PSNR | BPP | CR |
| 1 | 27040 | 3.811 | 0.0078 | 0.03 |
| 2 | 8088 | 9.052 | 0.0081 | 0.03 |
| 3 | 3170 | 13.12 | 0.0078 | 0.03 |
| 4 | 2489 | 14.17 | 0.0081 | 0.03 |
| 5 | 2289 | 14.53 | 0.0081 | 0.03 |
| 6 | 1774 | 15.64 | 0.0098 | 0.04 |
| 7 | 1217 | 17.28 | 0.0175 | 0.07 |
| 8 | 816.3 | 19.01 | 0.0358 | 0.15 |
| 9 | 536.4 | 20.84 | 0.0859 | 0.36 |
| 10 | 347.7 | 22.72 | 0.2132 | 0.89 |
| 11 | 188.9 | 25.37 | 0.6272 | 2.61 |
| 12 | 87.1 | 28.73 | 1.5223 | 6.34 |
| 13 | 36.12 | 32.55 | 3.024 | 12.60 |
| 14 | 13.56 | 36.81 | 5.339 | 22.25 |
| 15 | 4.954 | 41.18 | 8.3561 | 34.82 |
| 16 | 2.506 | 44.14 | 11.573 | 48.22 |
| 17 | 1.987 | 45.15 | 14.894 | 62.06 |

**Table 3: The compression after de nosing Image1 used FDSCWT and FDTCWT**

| Iterations | FDSCWT & FDTCWT | | | |
|---|---|---|---|---|
| | MSE | PSNR | BPP | CR |
| 1 | 11010 | 7.712 | 0.008 | 0.03 |
| 2 | 3654 | 12.50 | 0.008 | 0.03 |
| 3 | 3654 | 12.50 | 0.008 | 0.03 |
| 4 | 3094 | 13.23 | 0.008 | 0.03 |
| 5 | 2068 | 14.98 | 0.011 | 0.05 |
| 6 | 1669 | 15.91 | 0.016 | 0.07 |
| 7 | 1030 | 18.00 | 0.046 | 0.19 |
| 8 | 619.3 | 20.21 | 0.111 | 0.46 |
| 9 | 278.4 | 23.68 | 0.319 | 1.33 |
| 10 | 124.4 | 27.18 | 0.633 | 2.64 |
| 11 | 53.76 | 30.83 | 1.175 | 4.64 |
| 12 | 21.41 | 34.82 | 2.074 | 8.64 |
| 13 | 8.404 | 38.89 | 3.380 | 14.12 |
| 14 | 3.443 | 42.76 | 5.343 | 22.27 |
| 15 | 1.758 | 45.68 | 7.941 | 33.09 |
| 16 | 1.302 | 46.99 | 11.081 | 46.17 |
| 17 | 1.302 | 46.99 | 11.081 | 46.17 |

**Table 4: The compression after de nosing Image2 used FDSCWT and FDTCWT**

| | FDSCWT& FDTCWT | | | |
|---|---|---|---|---|
| Iterations | MSE | PSNR | BPP | CR |
| 1 | 26530 | 3.893 | 0.0078 | 0.03 |
| 2 | 7963 | 9.12 | 0.0081 | 0.03 |
| 3 | 3185 | 13.12 | 0.0078 | 0.03 |
| 4 | 2527 | 14.17 | 0.0081 | 0.03 |
| 5 | 2352 | 14.42 | 0.0081 | 0.03 |
| 6 | 1851 | 15.64 | 0.0098 | 0.04 |
| 7 | 1338 | 16.87 | 0.0177 | 0.07 |
| 8 | 948.1 | 18.36 | 0.0364 | 0.15 |
| 9 | 621.7 | 20.19 | 0.0954 | 0.40 |
| 10 | 334.5 | 22.89 | 0.2777 | 0.82 |
| 11 | 173.1 | 25.75 | 0.6760 | 6.12 |
| 12 | 80.61 | 29.07 | 1.4686 | 6.34 |
| 13 | 32.57 | 33 | 2.854 | 11.89 |
| 14 | 12.57 | 37.14 | 7.610 | 20.39 |
| 15 | 5.028 | 41.12 | 10.763 | 31.71 |
| **16** | 2.705 | 43.81 | 14.133 | 44.85 |
| **17** | 2.185 | 44.74 | 14.133 | 58.89 |

## 2.2. The Convolution Neural Network with Newproposed Filter

The neuron in the neural network is considered the main thing. Its work is similar to the work of the human brain in receiving instructions from the outside, which are called inputs Layers are created in the neural network through the connection point between the weight of the two neurons with a wave (1 + 1/2), which contributes to creating more than one layer until it reaches the output layers. When the waves are activated, neurons are built and deep learning is applied in the field of face recognition.

Mathematical aspects of convolutional neural network construction, starting from defining a new filter containing kernel N with the number of color image channels

Dimension of New Filter=$(L,L,N_c)$ (3)

Map the image with the convolutional neural network with the new filter

$$F(I,L)_{i,j} = \sum_{i=1}^{N_L} \sum_{j=1}^{N_W} \sum_{k=1}^{N_c} L_{i,j-L} I_{i+x-1,j+y-1,L} \quad (4)$$

$$Dim(W(I,L)) = \left(\left[\frac{N_L+2P-L}{S}\right]\right) \quad (5)$$

$$S = 1$$

$(N_L + 2P - L, N_W + 2P - L)$

The resulting grid must be as large as the output size P = (L-1) / 2 if L = 1 The face is identified or the image is identified while preserving the number of channels of the color image Figure (6) shows reducing the number of layers in a convolutional neural network with wavelet Figure (6) shows reducing the number of layers in a convolutional neural network with wavelet

$$Dim(W(I,L)) = \left(\left[\frac{N_L+2P-L}{S}+1\right], \left[\frac{N_W+2P-L}{S}+1\right], N_c\right) \quad S > 0 \quad (6)$$

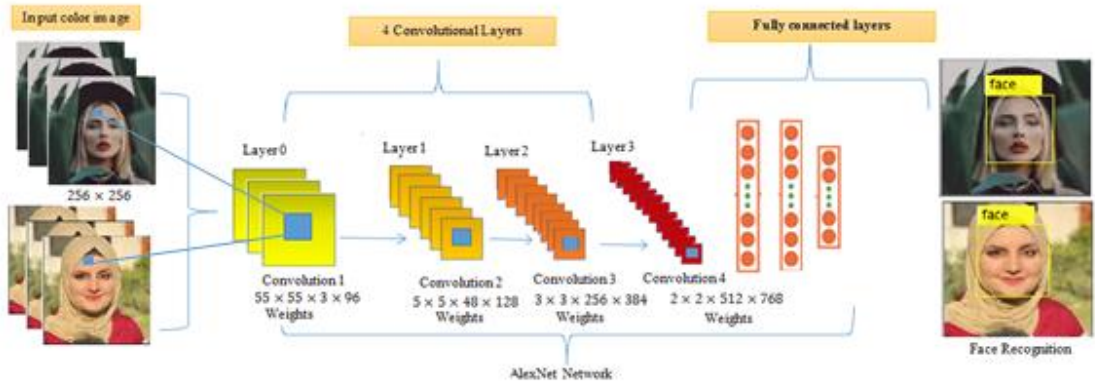

Figure 6. shows reducing the number of layers in a convolutional neural network with DSCWT and DTCWT

After applying the proposed new technique in face recognition, which is of great importance in the field of computer vision, using the MATLAB program, and to obtain the best results for measuring accuracy, using the proposed wavelets, with training the convolutional neural network with the CascadeObjectDetector function.

### 2.3. Efficient algorithm with DSCWCNN and DTCWCNN For Face Recognition

This algorithm to distinguish the face with the proposed new technique with the help of two separate waves of the second and third Chebyshev to analyse the image and raise the noise twice before compression and after compression after that the new filter is shed on the improved image with MATLAB program in training the convolutional neural network with the second type DSCWCNN and the same network with the third typeDTCWCNN, the following steps represent the algorithm.

In Put color Image

Step 1: Enhancement input image in Matlab program gets histogram the image
 details and the color layers that make up the image

Step 2: The layers of the input image are RGB using the new wavelets that are
 analysed into the aforementioned layers so that the image parameters are
 divided into approach parameters and the first details are concentrated in
 the first quadrant whose name is LL, but the details will be distributed in
 HL LH HH.

Step 3: Filters FDSCWTand FDTCWT The first noise from the input image is Removed.

Step 4: In this step, the image is compressed using the theory SPIHT in MATLAB
 program in true Compression to calculate the most important basics of
 image quality.

Step 5: Removing the noise from the image that was compressed in the previous step to improve the compressed image is called the second de noising.

Step 6: This step is the classification step with the two filters so that the convolutional neural network is created after programming the program in MATLAB by projecting the two filters on the image from which the noise has been removed twice on the first quarter of the analysed image and this stage is directed by

Vision.CascadeObjectDetector face_Detector = vision. CascadeObjectDetector

Out put Two images The face has been distinguished using the two proposed theories and the accuracy calculation shown by the Figure (7) shows the results of face recognition

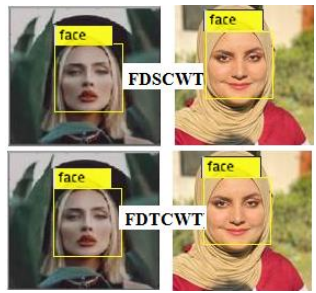

Figure7. shows the results of face recognition

## 3. DESCUTION RESULTS

Deep learning for face recognition after going through the preliminary stage in processing the face image, which includes noise augmentation and image compression, after building a new filter FDSCWT and FDTCW, to DSCWCNN and DTCWCNN, through the results reached in the tables above, the accuracy of the network was improved by analysing the image data entered in the classification process after calculating the most important criteria for image quality the Mean Square Error (MSE), Peak Signal to Noise Ratio (PSNR), Bit Per Pixel (BPP), and Compression Ratio (CR), and the problem of identifying faces was solved more accurately using the MATLAB program. Deep learning for face recognition after going through the preliminary stage in processing the face image, which includes noise augmentation and image compression, after building a new two filters FDSCWT and FDTCW And through the results reached in the tables above, the accuracy of the network was improved by analysing the image data entered in the classification process after calculating the most important criteria for image quality and the problem of identifying faces was solved more accurately using the MATLAB program. Figure (6) and (7) shows the quality of the accuracy of accuracy to reach the results 98.60% with DSCWCNN with time 1 min and 1 sec and 98.92 in 4 convolutions and 3 layers Pool 3×3 max pooling, 4 Relu1,2,3, and 4 with DTCWCNN with time 1 min and 19 sec in 4 of 4 Epoch while in other ways 6 of 6 Epoch after observing the tables that show the quality of the proposed method that was the reason for reaching these good results, as the error results started to go down to zero from the first step, which confirms the efficiency the proposed method Table (5) shows the privileges of the network trained in a very short time Table2: Convolutional neural network details with number of convolutions figure 8,9,10,11 represents the efficiency of the reads reached in tables 1,2,3,4 .

**Table 5: Convolutional neural network details with number of convolutions.**

| step | Name | Activations | Learnable |
|---|---|---|---|
| 1 | data $227 \times 227 \times 3$ image | $227 \times 227 \times 3$ | Weight $11 \times 11 \times 3 \times 96$ Bias $1 \times 1 \times 96$ |
| 2 | Conv1,1 69 $11 \times 11 \times 3$ image | $55 \times 55 \times 96$ | Weight $5 \times 5 \times 48 \times 128$ Bias $1 \times 1 \times 128 \times 2$ |
| 3 | Conv2, 2 groups of 128 image | $27 \times 27 \times 256$ | Weight $11 \times 11 \times 3 \times 96$ Bias $1 \times 1 \times 96$ |
| 4 | Conv3, 384 $3 \times 3 \times 256$ image | $13 \times 13 \times 384$ | Weight $3 \times 3 \times 256 \times 384$ Bias $1 \times 1 \times 384$ |
| 5 | Conv4, 512 $3 \times 3 \times 256$ image | $2 \times 2 \times 512$ | Weight $2 \times 2 \times 512 \times 768$ Bias $1 \times 1 \times 512$ |

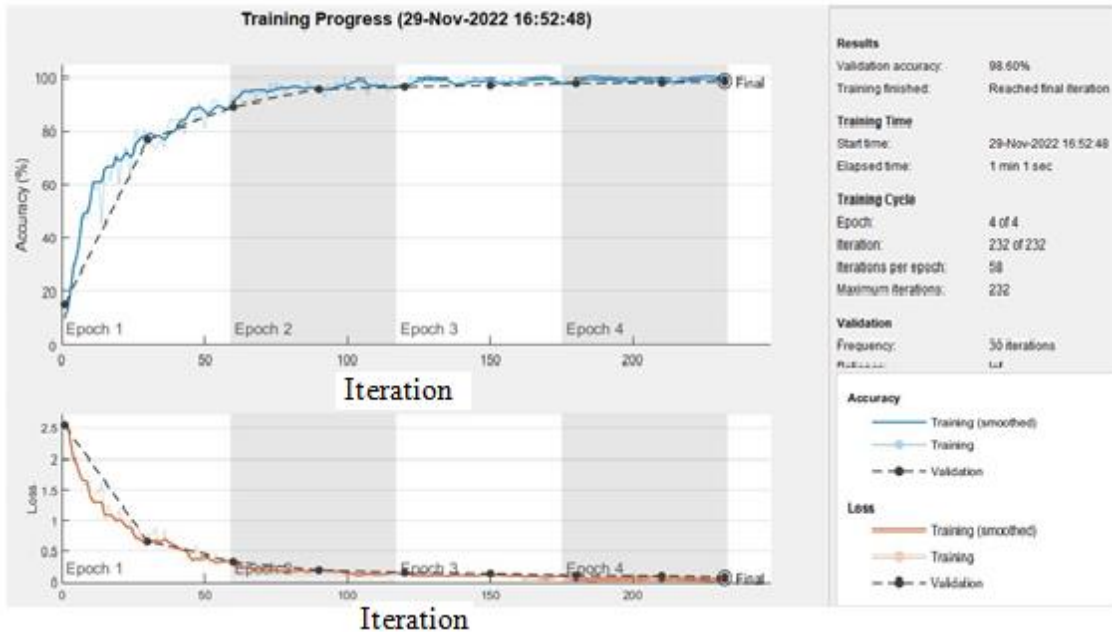

Figure 8. Shows the quality of the accuracy of accuracy with DSCWCNN

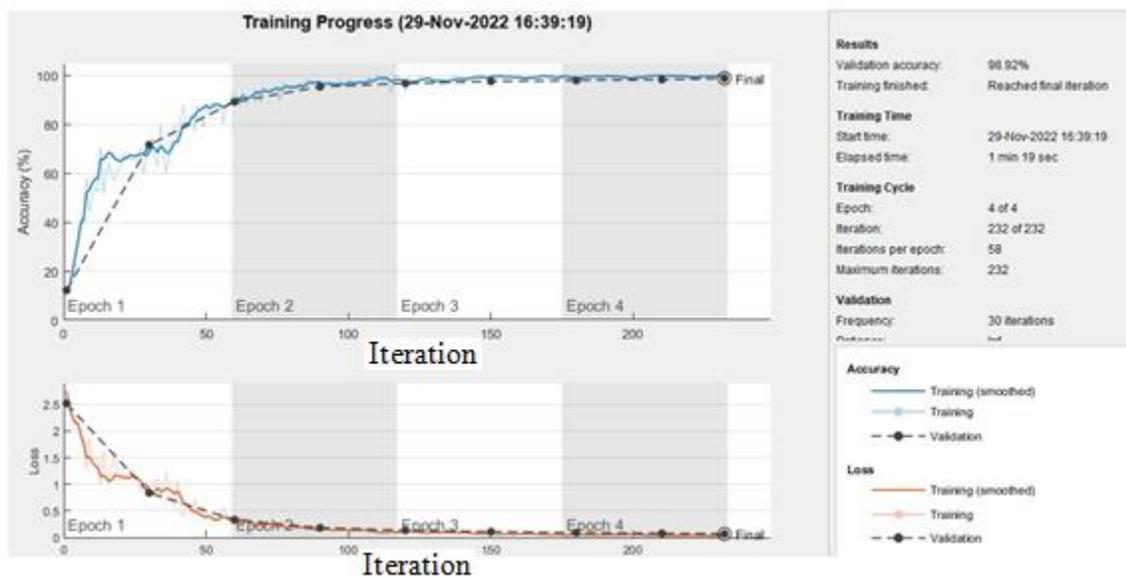

Figure 9. Shows the quality of the accuracy of accuracy with DTCWCNN

## 4. CONCLUSION

The discrete second Chebyshev wavelets transform (DSCWT) filter and the discrete third Chebyshev wavelets transform (FDTCWT filter) were proposed in this work. Discrete Third Chebyshev Wavelets Transform (FDTCWT) to perform the initial stage of processing, which proved its efficiency by using it to remove noise from the colour image, compress the image, reach good readings, and train the convolutional neural network based on the new wavelets the convolutional neural network they are Discrete Second Chebyshev Wavelet Convolution Neural Network (DSCWCNN) and Discrete Third Chebyshev Wavelet Convolution Neural Network (DTCWCNN) relying on the Alexnet network. with the Alexnet network with 4 convolutions to recognise faces The proposed algorithm is dubbed the "fast algorithm" because it achieves an accuracy of 98.60% with DSCWCNN in 1 minute and 1 second and 98.92% with DTCWCNN in 1 minute and 19 seconds.